\documentclass[letterpaper, 10 pt, journal, twoside]{ieeetran}
\usepackage{amsmath,amsfonts}
\usepackage{algorithmic}
\usepackage{array}
\usepackage[caption=false,font=normalsize,labelfont=sf,textfont=sf]{subfig}
\usepackage{textcomp}
\usepackage{stfloats}
\usepackage{url}
\usepackage{verbatim}
\usepackage{graphicx}

\usepackage{multirow}
\usepackage{pifont}
\usepackage{xcolor}
\newcommand{\cmark}{\ding{51}}%
\newcommand{\xmark}{\ding{55}}%
\usepackage{algorithm}

\def\BibTeX{{\rm B\kern-.05em{\sc i\kern-.025em b}\kern-.08em
T\kern-.1667em\lower.7ex\hbox{E}\kern-.125emX}}
\usepackage{balance}
\begin{document}
\title{EvDetMAV: Generalized MAV Detection from Moving Event Cameras}
\author{Yin Zhang$^{1,2}$, Zian Ning$^{2}$, Xiaoyu Zhang$^{3}$, Shiliang Guo$^{2}$, Peidong Liu$^{2}$, Shiyu Zhao$^{2}$
\thanks{
Manuscript received: March, 8, 2025; Revised May, 6, 2025; Accepted June, 21, 2025.}
\thanks{This paper was recommended for publication by Editor Giuseppe Loianno, upon evaluation of the Associate Editor and Reviewers’ comments. 
This research work was supported by the National Natural Science Foundation of China (Grant No. 62473320).
 (Corresponding author: Shiyu Zhao)}
 \thanks{$^{1}$Y. Zhang is with the College of Computer Science and Technology, Zhejiang University, Hangzhou, China {\tt\footnotesize zhangyin@westlake.edu.cn}}
 \thanks{$^{2}$Y. Zhang, Z. Ning, S. Guo, and S. Zhao are with the WINDY Lab, Department of Artificial Intelligence, Westlake University, Hangzhou, China. P. Liu is with the CVGL Lab {\tt\footnotesize\{ningzian, guoshiliang, liupeidong, zhaoshiyu\}@westlake.edu.cn}}
 \thanks{$^{3}$X. Zhang is with The Chinese University of Hong Kong, Hong Kong, China {\tt\footnotesize zhang.xy@link.cuhk.edu.hk}}
 \thanks{Digital Object Identifier (DOI): see top of this page.}
}
\markboth{IEEE ROBOTICS AND AUTOMATION LETTERS. PREPRINT VERSION. ACCEPTED June, 2025}%
{Zhang \MakeLowercase{\textit{et al.}}: EvDetMAV: Generalized MAV Detection from Moving Event Cameras}

\maketitle

\begin{abstract}
Existing micro aerial vehicle (MAV) detection methods mainly rely on the target's appearance features in RGB images, whose diversity makes it difficult to achieve generalized MAV detection. We notice that different types of MAVs share the same distinctive features in event streams due to their high-speed rotating propellers, which are hard to see in RGB images. This paper studies how to detect different types of MAVs from an event camera by fully exploiting the features of propellers in the original event stream. The proposed method consists of three modules to extract the salient and spatio-temporal features of the propellers while filtering out noise from background objects and camera motion. Since there are no existing event-based MAV datasets, we introduce a novel MAV dataset for the community. This is the first event-based MAV dataset comprising multiple scenarios and different types of MAVs. Without training, our method significantly outperforms state-of-the-art methods and can deal with challenging scenarios, achieving a precision rate of 83.0\% (+30.3\%) and a recall rate of 81.5\% (+36.4\%) on the proposed testing dataset. The dataset and code are available at: https://github.com/WindyLab/EvDetMAV.
\end{abstract}


\section{Introduction}
\IEEEPARstart{V}{ision}-based MAV detection plays an important role in various applications such as MAV-pursuing-MAV systems \cite{ning2024bearing}, see-and-avoid systems \cite{sapkota2016vision}, and MAV swarms \cite{wang2024tansformloc, schilling2021vision}. Most existing methods rely on RGB cameras to detect MAVs using their appearance and motion cues \cite{ashraf2021dogfight}. However, the diverse RGB appearances of MAVs make it difficult for visual algorithms to detect multiple types of MAVs in complex environments and achieve generalized MAV detection \cite{zhang2024domain}. 

\begin{figure}[t]
\centering
\includegraphics[width=1\columnwidth]{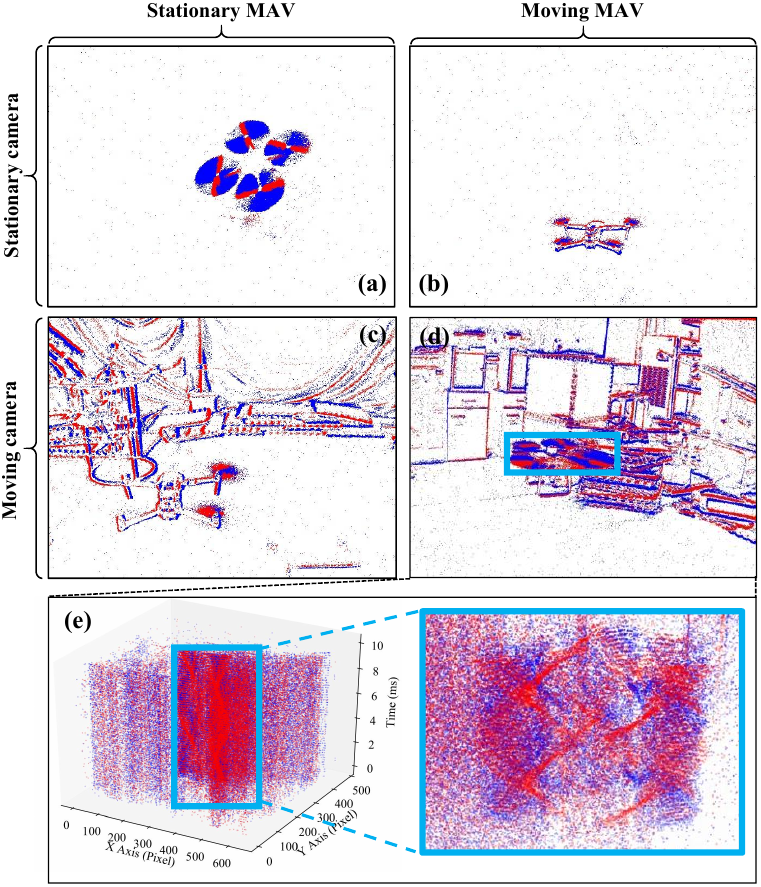} 
\caption{Four types of event images generated from different motion combinations of the camera and the MAV. (a) and (b) are captured by a stationary camera, while (c) and (d) involve a moving camera. (e) presents the original event stream, along with a zoomed-in view of the propellers in (d), highlighting their distinct motion patterns.}
\label{MainFigure}
\end{figure}

A common feature owned by different types of MAVs is that they all have fast-rotating propellers, which spin at speeds ranging from $5k$ to $15k$~RPM \cite{quan2017introduction}. Therefore, leveraging this feature can be key to detecting different types of MAVs. 
Event cameras, as neuromorphic vision sensors, offer distinct advantages for capturing such motion dynamics due to their high temporal resolution, low latency, and high dynamic range \cite{gallego2020event}. Unlike conventional RGB cameras, event cameras asynchronously record changes in pixel intensity and output a stream of events.
As shown in Fig.~\ref{MainFigure}(e), although these fast-rotating propellers are hard to see in the RGB images, they produce clear and distinctive patterns in the event stream that set them apart from other moving objects. The rotation of the propellers generates numerous positive and negative events around the propeller areas, which quickly form a spiral pattern, making the propellers distinguishable from other objects from any perspective. 

Recent studies have utilized event cameras for MAV detection, but the temporal and distinctive features of propellers have remained underutilized. The EvPropNet method \cite{sanket2021evpropnet} detects different types of MAVs by locating propeller centers in the event images. It works well under very slow motion when the camera and the target are close. The EV-Tach method \cite{zhao2023ev} detects MAVs by finding the grid with the highest grayscale value in the accumulated event map, but it is designed only for stationary scenarios, as shown in Fig.~\ref{MainFigure}(a). The work in \cite{iaboni2022event} detects fast-moving MAVs using a stationary event camera, as shown in  Fig.~\ref{MainFigure}(b). In summary, these approaches are suitable for simple scenarios but are susceptible to background noise interference.

\begin{figure*}[t]
\centering
\includegraphics[width=2\columnwidth]{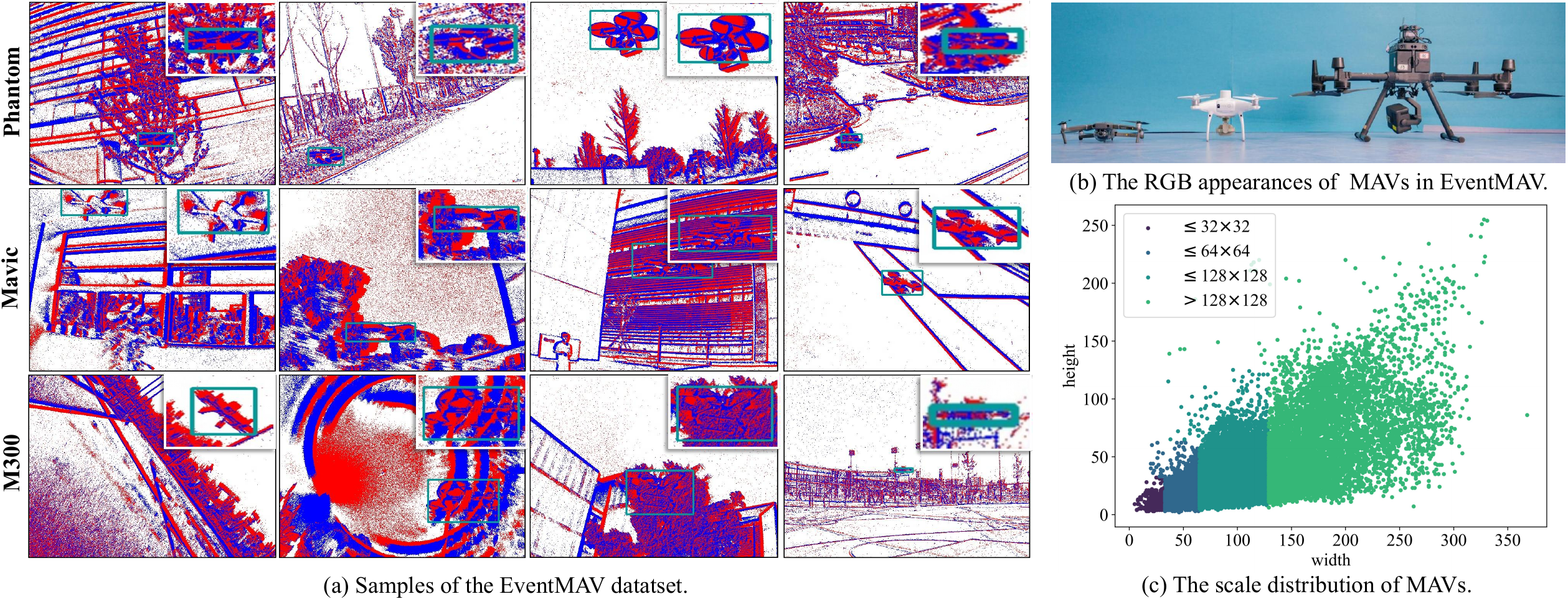} 
\caption{Visualization and statistics of the EventMAV dataset. The three rows of (a) contain three types of MAVs whose RGB appearances are shown in (b). The scale distribution of MAVs is shown in (c).}
\label{Dataset}
\end{figure*}
MAV detection from moving event cameras is a critical challenge, but has not been well studied so far. The appearances of MAVs in the event stream vary depending on the motions of the camera and the target. As shown in Fig.~\ref{MainFigure}, event data can be classified into four types. The MAV is visible and easy to detect when the camera is stationary. However, when the camera moves, background noise and other moving objects make the detection difficult. 
Despite such significant noise, the high-speed rotation of the propellers creates distinctive patterns in the event stream, making it possible to identify MAVs, as shown in Fig.~\ref{MainFigure}(e).
We focus on improving detection performance under these challenging scenarios.

In this paper, we aim to detect multiple types of MAVs by detecting their propellers in the event stream under complex conditions. Our method exploits the propellers' saliency and periodicity features that are not well explored and achieves generalized MAV detection without the need for training. The contributions of our work are as follows.

1) To address the challenges posed by camera motion, we propose a novel method, EvDetMAV, for detecting MAVs from the event stream. First, a saliency map is generated using a simple yet effective principle to enhance the distinctiveness of the propellers. Next, spatio-temporal features are extracted to further filter out background noise. Finally, a precise segmentation result is obtained through the clustering-based module.

2) We introduce a new event-based MAV dataset, EventMAV, which contains $25,335$ event periods, each with a corresponding bounding box, as shown in Fig.~\ref{Dataset}. To the best of our knowledge, this is the first dataset dedicated to event-based MAV detection. To guarantee its diversity, the data is collected from different perspectives, various scenarios, and three types of MAVs.

3) Extensive experiments demonstrate the effectiveness of our method. Without requiring training, our method achieves 83.0\% (+30.3\% over EvPropNet \cite{sanket2021evpropnet}) precision and 81.5\% (+36.4\%) recall on the EventMAV dataset, significantly outperforming existing state-of-the-art methods. This work demonstrates the feasibility of event cameras for generalized MAV detection.

\section{Related Work}

\subsection{MAV Detection from RGB Cameras}
RGB-based MAV detection methods primarily rely on appearance and motion information \cite{ashraf2021dogfight}. Appearance-based methods can be classified into traditional feature descriptor-based methods \cite{rozantsev2016detecting} and deep learning-based methods \cite{pavliv2021tracking, vrba2020marker}. Some works incorporate motion cues into their methods to combine the temporal information with the appearance features to achieve better detection performances \cite{guo2024global}. Even though the learning-based method performs well on the testing subset with the same distribution, the performance will drop sharply when meeting unseen MAVs and unknown environments due to the diversity of RGB images and the domain gap\cite{zhang2024domain}. Therefore, generalized MAV detection is still a challenge that needs to be solved.

\subsection{MAV Detection from Event Cameras}
There are a few works that adopt event cameras to detect MAVs. EvPropNet \cite{sanket2021evpropnet} proposes a synthetic data generation method to generate event images that contain propellers, but its effectiveness is limited to slow-moving cameras, close-range settings, and vertical-view scenarios. EV-Tach \cite{zhao2023ev} estimates the rotating speed of the propellers based on ICP registration \cite{besl1992method}, but its detection accuracy depends heavily on the initial clustering center, making it sensitive to background noise. The work in \cite{iaboni2022event} proposes detecting fast-moving MAVs from a stationary event camera using the YOLOv5 detection network \cite{yolov5}, which is a simple scenario shown in Fig.~\ref{MainFigure}(b) where the camera does not move. Overall, the above methods are only suitable for simple scenarios and do not fully exploit the distinctive features of propellers in event streams. 

\begin{figure*}[t]
	\centering
	\includegraphics[width=2\columnwidth]{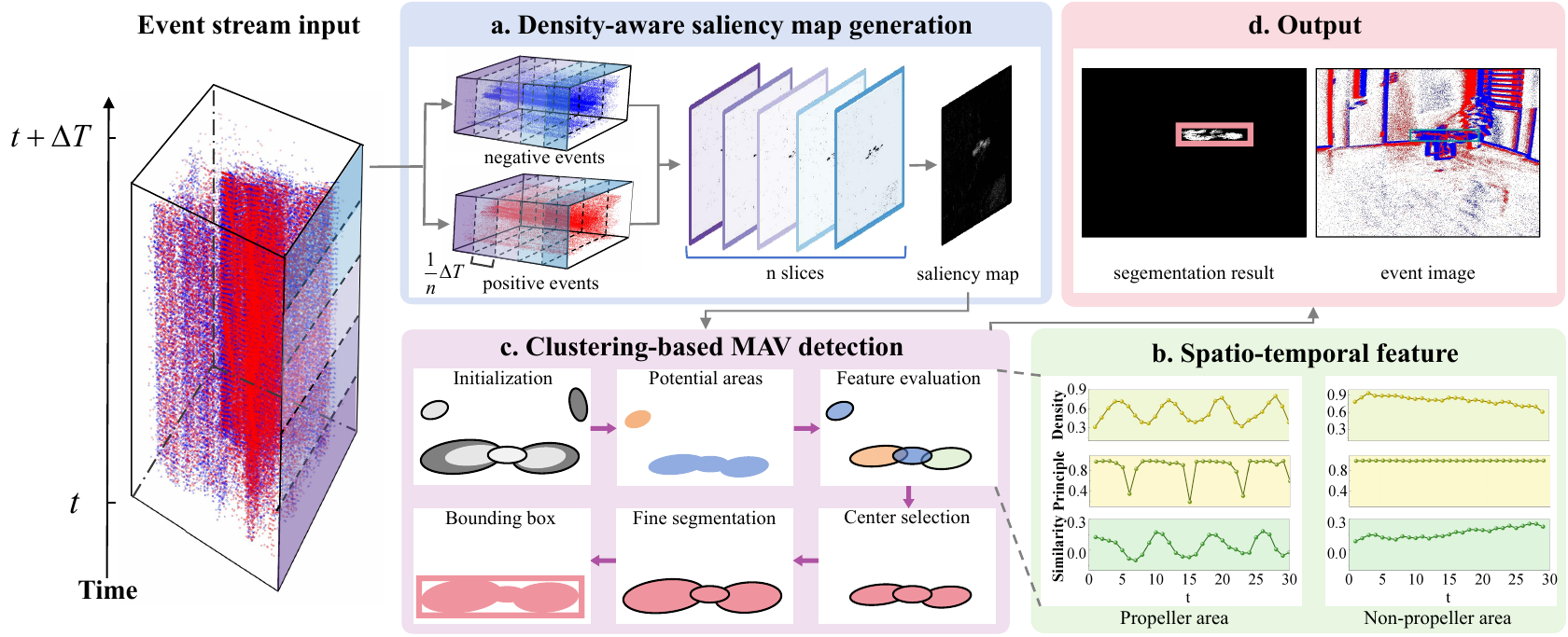} 
	\caption{The proposed method consists of three modules. The module in (a) generates a saliency map based on a density-aware principle. The module in (b) filters out excessive background noise. Finally, the coarse-to-fine module in (c) outputs the final bounding box of the target.}
	\label{Framework}
\end{figure*}

\subsection{Object Detection from Event Cameras}
The event representations used in event-based computer vision can be classified into two categories: point-based and image-based representations \cite{peng2023get}. Many state-of-the-art methods still choose event images as inputs to use the well-developed convolution neural networks (CNNs) as backbones \cite{cannici2019asynchronous}. The work in \cite{gehrig2023recurrent} proposes a recurrent transformer-based network for event-based object detection based on a series of slice images. However, this type of method could not fully utilize the temporal information. To utilize the sparse feature of the event camera, some works choose spiking neural networks (SNNs)  as backbones to achieve event-based object detection from event points \cite{cordone2022object, cao2024chasing}. Even though SSN is suitable for the event stream, its slow development of SNN technology makes the detection performance not comparable with CNN-based methods \cite{gehrig2024low}. In summary, integrating event data with deep learning models still meets challenges like data sparsity and computational efficiency.

\begin{table}[t] 
\centering
\setlength{\tabcolsep}{4pt}
\caption{Key attributes of the proposed dataset}
\begin{tabular}{c|c|c|c|c} 
\hline\hline
Attributes   & Environments & Target numbers & MAV types & Size range\\ \hline
Values & 15 & one per period & 3 & 0.37-0.81 m \\
  \hline \hline
\end{tabular}
\label{tab:attribute}
\end{table}

\begin{table}[t]
\centering
\caption{Details of different subsets in the EventMAV dataset}
\begin{tabular}{c|c|c|c|c}
\hline \hline
\textbf{MAV} & \textbf{Training Set} & \textbf{Validation Set} & \textbf{Testing Set} & \textbf{All}\\ \hline
 Phantom & 4,302 & 717 & 2,152 & \textbf{7,171}\\
 Mavic & 6,111 & 1,019 & 3,056 & \textbf{10,186} \\
 M300 & 4,786 & 798 & 2,394 & \textbf{7,978} \\\hline
 \textbf{All} & \textbf{15,199} & \textbf{2,534}& \textbf{7,602} & \textbf{25,335}\\\hline\hline 
\end{tabular}
\label{Subsets}
\end{table}

\subsection{Moving Object Detection from Event Cameras}
Some researchers specialize in moving object detection from event cameras \cite{liu2023motion}. The methods can be classified into traditional feature-based methods \cite{stoffregen2019event} and supervised learning-based methods. The 3D geometry of the event stream is used in \cite{mitrokhin2018event} for calculating the motion compensation matrix to detect moving objects. Evdodgenet \cite{sanket2020evdodgenet} adopts a network to detect moving objects. SpikeMS \cite{parameshwara2021spikems} proposes to employ SNNs to segment moving objects from the event stream. Even though MAVs are also moving objects, we focus on leveraging their distinctive features of propellers.

\section{Proposed Dataset}

This section presents EventMAV, the first event-based MAV dataset that includes multiple MAV types and diverse environments, all captured using a moving event camera.

The EventMAV dataset includes three types of MAVs: DJI Mavic, DJI Phantom, and DJI M300. Figure~\ref{Dataset} illustrates their RGB and event-based appearances. Key attributes of the dataset are summarized in Table~\ref{tab:attribute}. The dataset is captured using a DVXplorer Micro event camera with a resolution of $640\times480$. To ensure diverse and challenging detection scenarios, data collection spans various environments, perspectives, and distances. A total of 15 types of indoor and outdoor environments are included, such as hallways, parks, and tennis courts. Figure~\ref{Dataset} presents sample images and the scale distribution of targets, with each row corresponding to a different MAV type. 

The EventMAV dataset consists of $25335$ event periods, with the distribution across subsets detailed in Table~\ref{Subsets}. The time window for each period is categorized into four options: 10 ms, 15 ms, 20 ms, and 30 ms. Ground truth bounding boxes are manually annotated on the final event images for each period, with labels corresponding to the regions of the propellers. Each period contains one MAV target. Specifically, the EventMAV-Phantom, EventMAV-Mavic, and EventMAV-M300 subsets contain $7171$, $10186$, and $7978$ periods, respectively. 
The dataset is partitioned into training, validation, and testing subsets to establish a benchmark, facilitating future research and advancements in event-based MAV detection.

\section{Proposed Method}
To leverage the propeller features in the event stream, we propose EvDetMAV, a density-aware point-based method for MAV detection using an event camera. Our goal is to obtain the bounding box of the MAV in the event image from a short period of the event stream. As shown in Fig.~\ref{Framework}, the method consists of three modules. 1) The density-aware saliency map generation module rapidly extracts salient propeller features using a simple yet effective principle. 2) The spatio-temporal feature extraction module identifies potential propeller areas while reducing noise interference. 3) The clustering-based MAV detection module generates bounding boxes for the propeller areas. Details of these three modules are given below.

\begin{figure}[t]
	\centering
	\includegraphics[width=1\columnwidth]{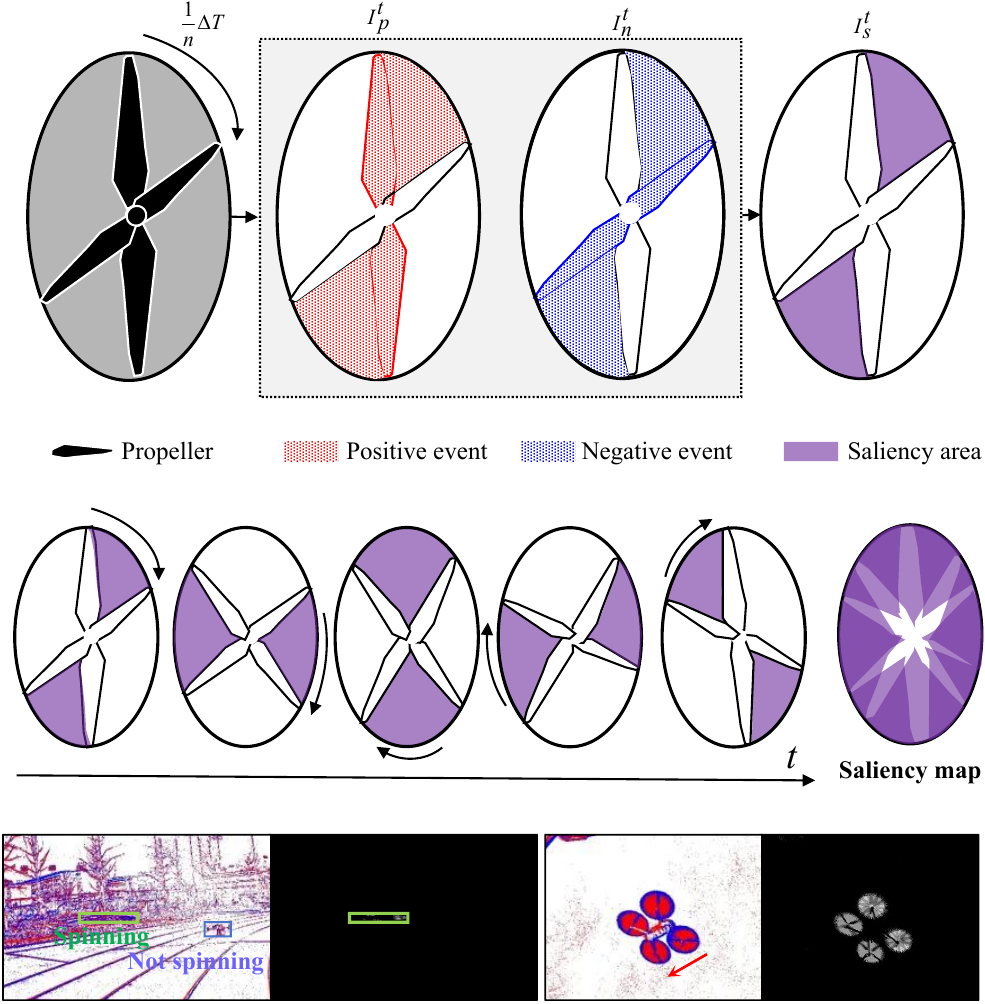} 
	\caption{The principle of the saliency map generation process. The bottom presents some results of this module.}
	\label{Saliency}
\end{figure}

\subsection{Density-Aware Saliency Map Generation} 
\label{sec:sal}
Each pixel in an event camera operates asynchronously and independently, generating an event when its logarithmic brightness change exceeds a threshold. Each event in the stream is denoted as $e(x, y, t, p)$, where $t$ is the timestamp, $(x, y)$ represents the pixel coordinates, and $p$ indicates the polarity. Our goal is to obtain the propeller bounding box $P(x, y, w, h)$ over a short period $\Delta T$. This period is divided into $n$ parts. Given that MAV propellers rotate at high speeds (e.g. $10k$~RPM), both positive and negative events are frequently triggered in the propeller region. We leverage this unique density distribution characteristic to generate a saliency map, enabling rapid identification of potential propeller regions.

First, a binary map is generated to identify the potential area of propellers over a short interval $\frac{1}{n}\Delta T$. As shown in Fig.~\ref{Saliency},  the propeller rotates by a certain angle within $\frac{1}{n}\Delta T$. The positive and negative events generate two event images  $I_p^t(i, j)$ and $I_n^t(i, j)$ during this period. The rotation of the propeller forms two triangular regions on $I_p^t(i, j)$ and $I_n^t(i, j)$. The intersection of these regions, $I_s^t(i, j) = I_p^t(i, j) \cap I_n^t(i, j)$, contains both positive and negative events, making it a distinctive feature that differentiates propellers from other objects. This region consistently generates the same events when the propeller reappears in subsequent rotation cycles.

Then, over a period $\Delta T$, the saliency map $I_s$ is obtained by accumulating a series of $I_s^t$ where $t\in[1, n]$. This process effectively suppresses background noise caused by camera motion and other moving objects. The bottom of Fig.~\ref{Saliency} presents saliency map results. In the left image, a moving MAV is in the air, while a stationary MAV remains on the ground. The propellers of the stationary MAV are not spinning, and thereby the target is not detected. Our method enhances the saliency of the moving MAV’s propellers while suppressing irrelevant noise. The right image shows a MAV moving in one direction, where the bottom propellers rotate faster than the top ones. Consequently, the four propellers exhibit different saliency values due to their varying rotation speeds. 

In summary, unlike existing saliency maps that directly accumulate events \cite{zhao2023ev}, the proposed module leverages the distinctive features of the propellers and effectively filters out background noise. More saliency map results are presented in Fig.~\ref{Comparison} to further demonstrate the effectiveness of this module.

\subsection{Spatio-Temporal Feature Extraction}
\label{sec:period}
In this section, we introduce a spatio-temporal feature extraction module that exploits the propellers' features to enhance the identification of propeller regions.

\subsubsection{Saliency scores} 

Regions with high saliency values are more likely to correspond to propeller areas. The saliency score $s_s$ is determined by the grayscale value and the area.

\subsubsection{Periodicity scores}
A local event stream can be extracted using the saliency map. As shown in Fig.~\ref{Framework}, the event stream from a propeller differs significantly from that of other objects. The event stream is divided into $m$
slices over a shorter time interval, generating a sequence of local images. Several feature descriptors are then applied to determine whether a given local region corresponds to a propeller.

\textbf{Density}. When a region contains a propeller, its event density fluctuates more rapidly than in other areas. Thus, density serves as a key feature descriptor. We define the density $f_d$ as the number of positive events in each slice.

\textbf{Structural similarity}. The structure of a slice changes rapidly when it contains propellers. The structural similarity $f_s$ is computed using the cosine similarity between two normalized one-dimensional vectors $\bar{x}_i$ and $\bar{x}_j$, derived from consecutive slice images. The vectors are normalized by subtracting their mean and dividing by their standard deviation. The cosine similarity is then calculated as $f_s = \bar{x}_i^T\bar{x}_j$.

\textbf{Principal direction similarity}. As a self-rotating object, the direction of the propeller changes rapidly across different slices. Even though the event points of different timestamps are compressed into one slice image, we can also extract the 2D points  $\mathcal{P}=[p_1,p_2, ..., p_w]^T \in \mathbb{R}^{w\times2}$ from the slice where  $p_i=(p_i^x, p_i^y) \in \mathbb{R}^{1\times 2}$ is the image coordinate and $w$ is the number of 2D points. The center of $\mathcal{P}$ is represented as $p_c=\frac{1}{w}\sum p_i$.
Then, the covariance matrix $C$ can be calculated by
\begin{equation}
    C = \frac{(\mathcal{P}-p_c \otimes \mathbb{I}_{w})^T(\mathcal{P}-p_c \otimes \mathbb{I}_{w})}{w},
\end{equation}
where $\mathbb{I}_{w} \in \mathbb{R}^{ {w}\times 1}$ is a vector of ones.
The matrix $C$ is a second-order real symmetric matrix with two eigenvalues. The largest eigenvalue $\lambda_{max}$ and its corresponding eigenvector $\xi$ has the relation $\lambda_{max} C = \xi C$.
$\xi$ represents the principal direction of the event points. Then, the angle between two vectors $\xi_1$ and $\xi_2$ of two consecutive  slice images can be calculated as
\begin{equation}
   f_p = \frac{|\xi_1^T\xi_2|}{||\xi_1||||\xi_2||}.
\end{equation}

Compared to other objects in the event stream, the propeller region exhibits periodic features. The changes in the three features $f_d$, $f_s$, and $f_p$ differ from those in non-propeller areas. The peaks and valleys are calculated after moving average filtering. $s_p$ can be calculated by whether $f_d$, $f_s$, and $f_p$ have peaks and valleys, and thereby has the maximum value of 6. The correlation of the periodic data has distinctive peaks and valleys compared with other non-periodic data, as shown in Fig.~\ref{Framework}(b). Areas with higher scores are more likely to correspond to the propeller region.

\subsection{Clustering-based MAV Detection}
Based on the saliency map, we introduce a clustering-based module to detect the final propeller areas of the MAVs. A coarse-to-fine strategy is employed to refine the detection of the propeller areas.

\begin{algorithm}[htbp]
	\caption{Clustering-based MAV Detection}
	\label{Cluster}
	\textbf{Input}:Event stream $e_i(x, y, t, p)$, $t \in [1, \Delta T]$ \\
	\textbf{Output}: Propeller area $P_i(x, y, w, h)$
	
	\begin{algorithmic}[1]
		\STATE \textbf{Initialization:} Obtain potential areas $A_i (i \in[1, N])$. \\
		\STATE \textbf{Coarse stage:} 
        \FOR {each $i\in [1,N]$}
         \STATE Calculate the saliency scores $s_s$ from Section~\ref{sec:period};  \\
        \ENDFOR
        \FOR {each $i\in [1,K]$}
         \STATE Calculate the periodicity scores $s_p$ from Section~\ref{sec:period};  \\
        \textbf{if} {$s_p \geq \tau_p$:} \textbf{then} Update $P_i$; \textbf{end if}
        \ENDFOR
		\STATE \textbf{Fine stage:} 
        \FOR {each $i \in [1,n_p]$}
        \STATE Calculate the Gaussian shape; \\
        \textbf{if} {consistent with $P_i$:} \textbf{then} update $P_i$; 
        \ENDFOR
		\STATE \textbf{return} $P_i$
	\end{algorithmic}
\end{algorithm}

\begin{figure}[t]
	\centering
	\includegraphics[width=1\columnwidth]{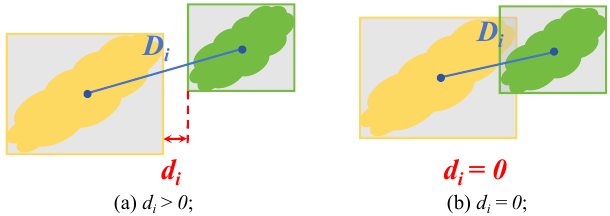} 
	\caption{The illustration of the clustering distance principles of $d_i$ and $D_i$. $D_i$ represents the Euclidean distance and $d_i$ shows the minimum distance.}
	\label{Clustering}
\end{figure}

1) \emph{Initialization}:
The saliency map is divided into potential areas based on predefined clustering rules. First, a threshold $\tau_s$ is applied to the saliency map $I_s$ to obtain saliency maps with only high values. The centers of the connected regions are then selected as initial clustering centers.  As shown in Fig.~\ref{Clustering}, instead of using the Euclidean distance $D_i$ between two center points, we use the minimum distance $d_i$ between two rectangle regions as the clustering distance. The clustering areas are updated throughout the process. Some clustering results are shown in Fig.~\ref{ablationstudy}.

2) \emph{Coarse stage}:
 First, the saliency scores of each area are calculated as described in Section~\ref{sec:sal}. The top $K$ areas are then selected for further processing. Finally, the area with the highest periodicity score that exceeds the threshold $\tau_p$ is identified as the propeller area. In this way, the coarse area of the propellers is obtained.

3) \emph{Fine stage}:
To achieve a more precise segmentation, the connected region from $I_s$ will be regarded as the result of the propeller area when its shape remains consistent. This ensures that background noise within the propeller area is excluded from the final detection result.

As shown in Fig.~\ref{Framework}(d), the clustering module ultimately provides a precise segmentation result along with the corresponding bounding box, ensuring accurate MAV detection.

\section{Experiments}

\subsection{Implementation Details}

We choose Precision Rate, Recall Rate, F1-Score, and mAP (mean Average Precision) as the evaluation metrics. The bounding box encompassing the propeller area is labeled as the ground truth. For the proposed method, the parameters $\tau_s$ and $\tau_p$ are set to $50$ and $3$, respectively. The IoU (Intersection-Over-Union) threshold is set to $0.4$. 

EV-Tach \cite{zhao2023ev} and EvPropNet \cite{sanket2021evpropnet} are chosen as comparison methods. The parameters for EV-Tach follow the settings outlined in the original paper. Since EvPropNet outputs keypoints of the propellers, we use the connected areas containing these keypoints as the final detection results to ensure a fair comparison. For the learning-based methods, we adopt the YOLOv5 network as used in \cite{iaboni2022event} for comparison.

\begin{figure*}[t]
	\centering
	\includegraphics[width=2\columnwidth]{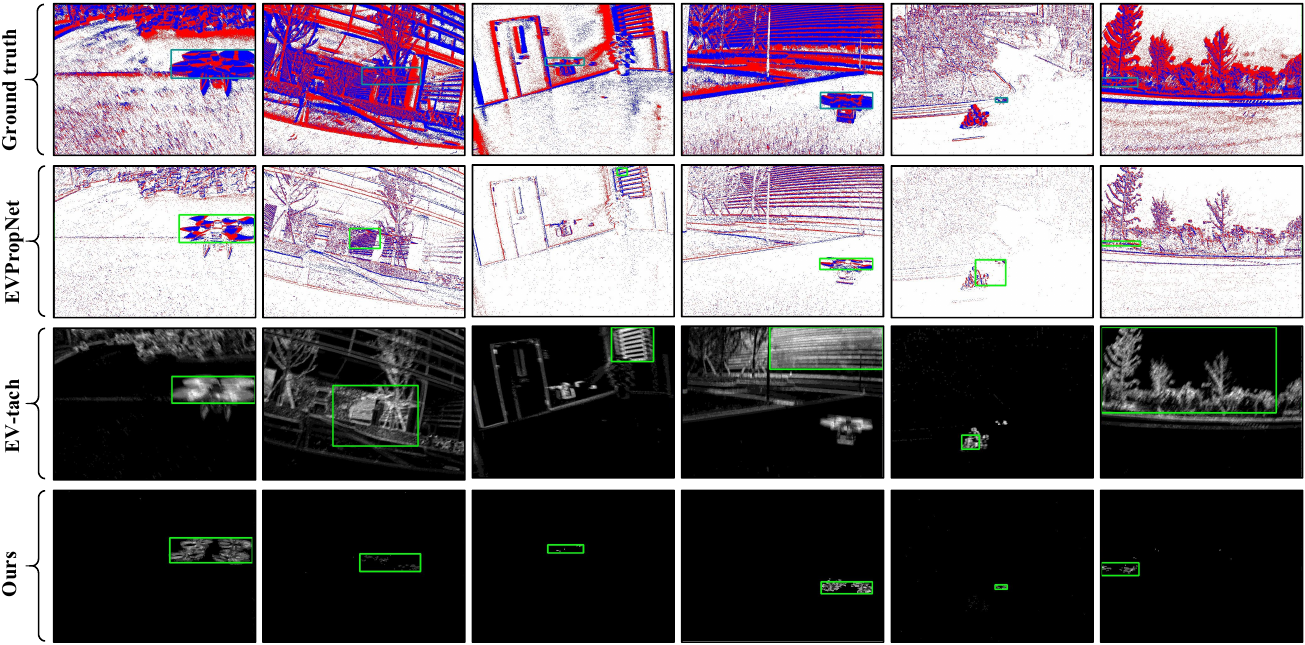} 
	\caption{The qualitative analysis of three methods. The event images of EvPropNet contain a shorter period for detection than the other two methods.}
	\label{Comparison}
\end{figure*}

\begin{table*}[htbp]
\caption{The detection results of different methods on the three testing subsets of EventMAV}
\centering
\setlength{\tabcolsep}{5pt}
\begin{tabular}{c|c|cc|cc|cc|ccc}
\hline\hline
\multirow{2}{*}{\textbf{Method}} &\multirow{2}{*}{\textbf{Need training?}} &  \multicolumn{2}{c|}{\textbf{EventMAV-Phantom}} &  \multicolumn{2}{c|}{\textbf{EventMAV-Mavic}} &  \multicolumn{2}{c|}{\textbf{EventMAV-M300}} &  \multicolumn{3}{c}{\textbf{EventMAV-All}}\\
\cline{3-11}
 & & \textbf{Precision} & \textbf{Recall}  & \textbf{Precision} & \textbf{Recall} &\textbf{Precision} & \textbf{Recall} &\textbf{Precision} & \textbf{Recall} & \textbf{F1}\\ \hline 
YOLOv5 \cite{yolov5, iaboni2022event}  & \cmark~(Phantom) & \textbf{94.8\%} & \textbf{96.6\%} & 77.6\% & 58.7\%  & 72.6\% & 51.6\%  & 82.3\% & 67.2\% & 74.0\% \\ \hline 
YOLOv5 \cite{yolov5, iaboni2022event} & \cmark~(Other two subsets) & 87.8\% & 77.9\%  &  \textbf{85.6\%} & 69.8\% &  76.7\% & 71.1\%  & \textbf{ 83.2\%} & 72.5\% & 77.5\% \\ \hline \hline
EvPropNet \cite{sanket2021evpropnet} &  \xmark & 54.9\% & 48.6\%   & 54.6\%& 50.6\% & 47.3\%  &35.0\% & 52.7\% & 45.1\% & 48.6\% \\ \hline
EV-Tach \cite{zhao2023ev}  & \xmark & 46.3\%& 46.3\%  &  46.8\% & 46.8\% & 40.7\% & 40.7\% & 44.8\% & 44.7\% & 44.7\%  \\ \hline

EvDetMAV (Ours)  &\xmark &85.0\% &82.9\% & 82.1\% & \textbf{80.6\%}  & \textbf{82.4\%} & \textbf{81.3\%} & 83.0\% & \textbf{81.5\%} & \textbf{82.2\%} \\ \hline \hline                         
\end{tabular}
\label{EResult}
\end{table*}  

\subsection{Experiments on the Proposed EventMAV Dataset}
We conduct experiments on the EventMAV testing set to evaluate the performance of different methods across different types of MAVs. The detection results are shown in Table~\ref{EResult}. Compared to methods that do not require training, our method achieves superior performance across all three subsets. The EvPropNet method \cite{sanket2021evpropnet} faces challenges when meeting small MAVs because the propellers are not obvious at greater distances. 
Additionally, the EV-Tach method \cite{zhao2023ev} is easily affected by background noise interference when the camera moves, resulting in inaccurate detection results. Even though the EventMAV dataset contains challenging scenarios like small targets, different perspectives, and complex camera motions, our method still performs well without the need for training. Our method achieves 83.0\% precision rate (+30.3\% over EvPropNet) and 81.5\% recall rate (+34.6\%). The proposed method significantly improves MAV detection under challenging scenarios by utilizing the unique features of propellers.

The proposed method presents greater potential for generalized MAV detection compared with existing deep learning-based methods. We evaluate the generalization ability of the YOLO network using two different training approaches. First, when the YOLO network is trained on the Phantom subset, its performance on the Phantom testing set is the best because the training and testing sets share the same distribution. However, its performance on the other two MAV types is lower than the Phantom and also lower than our method. Second, when the YOLO model is trained on two subsets and tested on the remaining one, the model benefits from more diverse training data, which improves its generalization ability. As a result, this approach achieves better performance than the first approach. Nevertheless, the final F1-Score of the YOLO model still falls short of our method. This is due to the proposed method’s ability to exploit the unique features of MAVs in event cameras. However, CNN-based methods cannot fully utilize the temporal information of event cameras.  We believe that combining our approach with deep learning methods could further enhance detection performance.

We also test our method on different types of MAVs. The detection results are shown in Table~\ref{wholesubsetresult}. It is important to emphasize that the experiments are conducted on the entire dataset, not just the testing set. The proposed method achieves 83.3\% precision rate and 80.1\% recall rate across these subsets without any training requirement. The performance demonstrates our method’s generalization ability for MAV detection, as it exploits the shared features of different MAV types in the event camera by extracting the spatio-temporal features of the propellers.

The average computation time of the proposed method on the EventMAV-Phantom subset is $0.02$ seconds on average on an i9-13980HX 2.40GHz CPU, which inherently has limited computational resources compared to a GPU. The average duration of event data is around 20 ms, which corresponds to real-time detection. 

\begin{figure}[t]
	\centering
	\includegraphics[width=1\columnwidth]{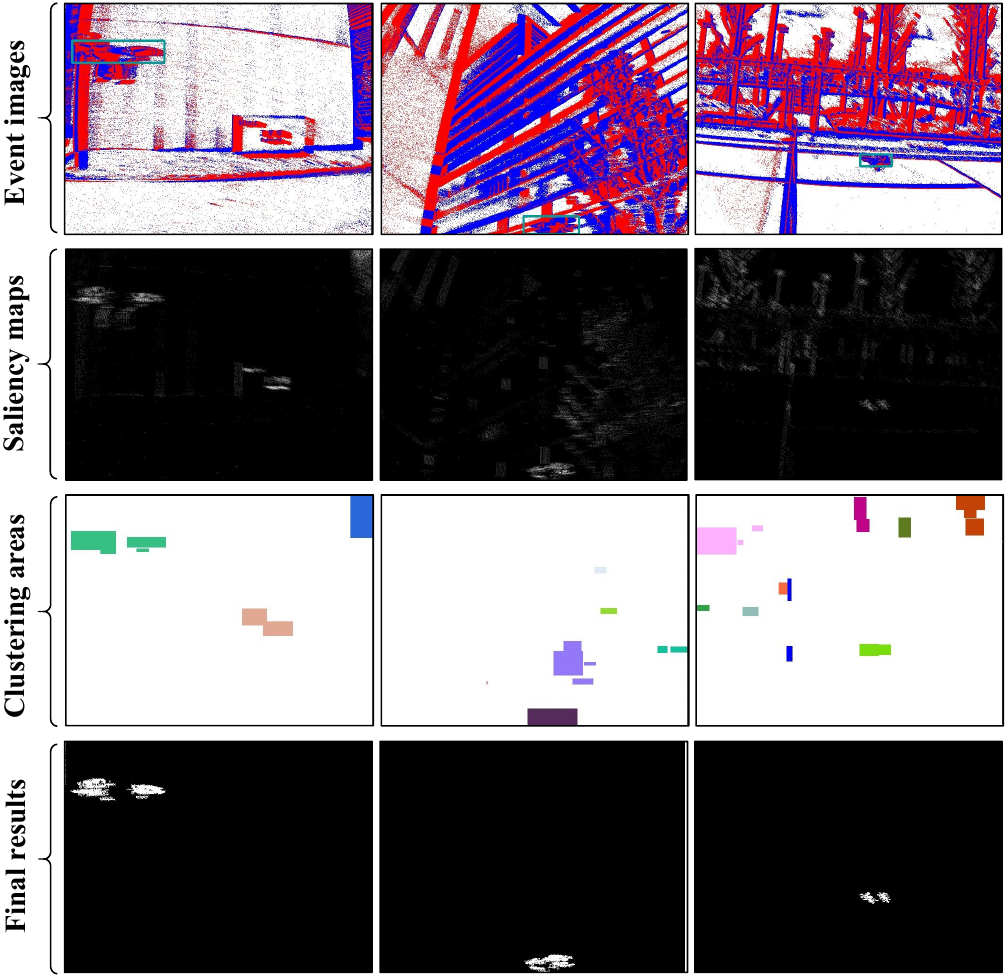} 
	\caption{Visualization of the outputs from each module of our method. 
    }
	\label{ablationstudy}
\end{figure}

\begin{table}[t]
\caption{Detection results on the whole EventMAV dataset}
\centering
\begin{tabular}{c|c|c|c}
\hline \hline
\textbf{Subsets} & \textbf{Precision Rate} & \textbf{Recall Rate} & \textbf{mAP} \\ \hline
EventMAV-Phantom & 83.9\% & 81.8\% & 69.4\% \\
 EventMAV-Mavic & 82.8\% & 81.0\% & 71.9\% \\
 EventMAV-M300 & 83.5\% & 77.5\% & 65.9\%\\\hline
 EventMAV & \textbf{83.3\%} & \textbf{80.1\%} & \textbf{67.4\%} \\ \hline \hline
\end{tabular}
\label{wholesubsetresult}
\end{table}
\subsection{Qualitative Analysis}

This section presents the qualitative analysis of the proposed method. Figure~\ref{Comparison} visualizes the detection results of different methods. The EvPropNet  \cite{sanket2021evpropnet} performs best when the camera has slow motion and a close view, but struggles in complex scenarios with rapid motion. The EV-Tach method \cite{zhao2023ev} is more susceptible to background noise, leading to incorrect initial clustering centers, which significantly impacts detection accuracy. Additionally, existing methods often fail to fully exploit the unique characteristics of propellers in event streams, limiting their robustness in challenging conditions. In contrast, our method effectively leverages the distinctive saliency and periodicity features of the propellers while suppressing background noise. As a result, it achieves more reliable and accurate detections across diverse environments and challenging motion conditions.

\subsection{Robustness Analysis}

We evaluate the robustness of our method by testing its performance under different conditions. The EventMAV-Phantom subset is chosen as the testing set. Table~\ref{robust} presents the performance of the proposed method across different scenarios, demonstrating its robustness and stability.

\subsubsection{Influence of illumination}
Illumination differences affect the number of events triggered in the propeller area. Indoor environments generate fewer events than outdoor settings, where higher illumination leads to increased event activity. Consequently, our method achieves better detection performance outdoors, reaching 72.1\% mAP, outperforming its indoor results.
\subsubsection{Influence of aspect ratio}
Viewing angle changes impact the aspect ratios of propellers, thereby affecting detection difficulty. This section examines the performance of our method at different aspect ratios of the targets. We divide the dataset into three subsets based on two thresholds. The detection results for these subsets are 62.6\%, 76.7\%, and 77.8\%, respectively. MAVs with smaller aspect ratios are the most challenging to detect, as the propellers occupy smaller areas under these conditions.
\subsubsection{Influence of scale}
The scale of MAVs in images affects detection difficulty. We divide the dataset into four subsets using the thresholds $32\times32$, $64\times64$, and $128\times128$. As the MAV scale increases, detection performance improves. However, when testing with smaller targets, detection performance drops to 39.7\% because the propellers are harder to distinguish at a distance. To enhance the detection of smaller MAVs, the algorithm should integrate additional information to handle these cases more effectively.

\begin{table}[t]
\caption{The robustness experiments on the proposed dataset}
\renewcommand{\arraystretch}{0.4}
\setlength{\tabcolsep}{5pt}
\centering
\begin{tabular}{c|c||c|c||c|c}
\hline \hline
\rule{0pt}{7pt} \textbf{Illumination} & \textbf{mAP} & \textbf{Aspect ratio
} & \textbf{mAP} & \textbf{Scale} & \textbf{mAP}\\ \hline 
\multirow{6}{*}{Indoor} & \multirow{6}{*}{57.0\%} & \multirow{4}{*}{ $\leq$0.3}& \multirow{4}{*}{62.6\%} & \multirow{3}{*}{Tiny} & \multirow{3}{*}{39.7\%} \\ 
 & & & & & \\ 
& & & & & \\ \cline{5-6}
& & & & \multirow{3}{*}{Small} & \multirow{3}{*}{68.2\%} \\ \cline{3-4}
& &\multirow{4}{*}{$\leq$0.6} & \multirow{4}{*}{76.7\%} &  & \\ 
& & & & & \\ \cline{1-2} \cline{5-6}
\multirow{6}{*}{Outdoor} & \multirow{6}{*}{72.1\%} & & &  \multirow{3}{*}{Medium} &  \multirow{3}{*}{74.8\%}\\ 
& & & & & \\  \cline{3-4}
& &\multirow{4}{*}{$>$0.6} & \multirow{4}{*}{77.8\%}& & \\ \cline{5-6}
& & & &  \multirow{3}{*}{Large}&  \multirow{3}{*}{81.4\%}\\ 
& & & & & \\ 
& & & & & \\ \hline  \hline
\end{tabular}
\label{robust}
\end{table}

\begin{table}[t]
\caption{The ablation study of the proposed method}
\centering
\begin{tabular}{l|c|c|c}
\hline \hline
\textbf{Method} & \textbf{Presion} & \textbf{Recall} &\textbf{mAP } \\ \hline
 Sal. (saliency) & 10.5\% &  33.1\% & 3.6\% \\
 Sal. + Extra. (feature extraction)  &31.6\% & 26.9\% & 8.7\% \\
 Sal. + Clus. (clustering) & 54.9\%& 76.7\% & 42.9\% \\\hline
Sal. + Extra. + Clus. & \textbf{85.0\%} &\textbf{82.9\%} & \textbf{70.8\%}  \\\hline\hline          
\end{tabular}
\label{ablation}
\end{table}

In summary, MAVs are more easily detected under strong illumination and with a close, vertical view.

\subsection{Ablation Study and Modular Analysis}
This section analyzes the impact of each module in the proposed method. Intermediate results are presented in Fig.~\ref{ablationstudy}. The first row shows the ground truth of the propeller areas. The second row shows the identified potential target regions. The third row illustrates the clustering results, with different colors randomly assigned for better visualization. The final row presents the refined segmentation results.

Table~\ref{ablation} presents the detection results on the EventMAV-Phantom subset, comparing different module combinations. The saliency module identifies potential connected regions of the targets, but it still includes some background areas. Regions corresponding to the propellers are treated as false negatives if their IoU values do not meet the threshold. The feature extraction module helps to filter out more noise and improves detection accuracy, but it still faces the IoU issue. Therefore, there is a significant improvement when the clustering module participates.

\subsection{Discussions}

One key limitation of the proposed method is that performance degrades when the target MAV is too far from the camera, as the propellers become more difficult to distinguish in the event stream. Additionally, the presence of a protective box or cover surrounding the propellers further complicates MAV detection.

The choice of $K$ is influenced by the number of target MAVs. Since there is one MAV per period in our dataset, $K$ is set to 4 for our experiments. $K$ can also be set as a dynamic variable according to the requirements. $K$ represents the initial seeds for the algorithm. If $K$ is set too large, the computation time increases, while if $K$ is set too small, there is a risk that the propeller area may not be selected as an initial seed.
 
\section{Conclusion}
This paper investigates generalized MAV detection using an event camera by exploiting the distinctive features of fast-rotating propellers. Unlike RGB cameras, event cameras offer unique advantages for this task. To address challenges posed by camera motion, we propose a point-based density-aware method for effectively extracting the saliency and spatio-temporal features of the propellers. In addition, we also introduce the first event-based MAV detection dataset that contains multiple types of MAVs and diverse scenarios for the community. By exploiting propeller features in event streams, our method demonstrates the feasibility of generalized MAV detection under challenging scenarios. Future work will integrate these features into neural networks to further enhance performance. We will also explore datasets involving multi-target MAVs.

\bibliography{zy} 
\bibliographystyle{ieeetr}

\end{document}